# Ensemble Learning Model on Artificial Neural Network-Backpropagation (ANN-BP) Architecture for Coal Pillar Stability Classification

Gabriella Aileen Mendrofa[1, a)], Bevina Desjwiandara Handari[1, b)], Gatot Fatwanto Hertono[1, c)]

[1]*Department of Mathematics, Faculty of Mathematics and Natural Sciences (FMIPA), Universitas Indonesia, Depok 16424, Indonesia*

a)gabriellaaileen07@gmail.com
b)bevina@sci.ui.ac.id
c)Corresponding author: gatot-f1@ui.ac.id

**Abstract.** Pillars are important structural units used to ensure mining safety in underground hard rock mines. Unstable pillars can significantly increase worker safety hazards and sudden roof collapse. Therefore, precise predictions regarding the stability of underground pillars are required. One common index that is often used to assess pillar stability is the Safety Factor (SF). Unfortunately, such crisp boundaries in pillar stability assessment using SF are unreliable. This paper presents a novel application of Artificial Neural Network-Backpropagation (ANN-BP) and Deep Ensemble Learning for pillar stability classification. There are three types of ANN-BP used for the classification of pillar stability distinguished by their activation functions: ANN-BP ReLU, ANN-BP ELU, and ANN-BP GELU. These three activation functions were chosen because they can solve the vanishing gradient problem in ANN-BP. In addition, a Deep Ensemble Learning process was carried out on these three types of ANN-BP to reduce the prediction variance and improve the classification results. This study also presents new labeling for pillar stability by considering its suitability with the SF. Thus, pillar stability is expanded into four categories: failed with a suitable safety factor, intact with a suitable safety factor, failed without a suitable safety factor, and intact without a suitable safety factor. There are five features used for each model: pillar width, mining height, bord width, depth to floor, and ratio. In constructing the model, the initial dataset is divided into training data, validation data, and testing data. In this case, four type of proportions are used. For training-testing division the proportions are: 80%:20%, 70%:30%, for training-validation-testing division the proportions are: 80%:10%:10%, 70%:15%:15%. Average accuracy, $F_1$-score, and $F_2$-score from 10 trials were used as performance indicators for each model. The results showed that the ANN-BP model with Ensemble Learning could improve ANN-BP performance with an average accuracy 87.98% and an $F_2$-score 99.27% for the category of failed with a suitable safety factor.

**Keywords:** *Artificial Neural Network, Ensemble Learning, Pillar Stability*

## INTRODUCTION

Pillars are important structural components in underground hard rock and coal mining. This is because pillars can provide temporary or permanent support for mining and tunneling operations [1]. Pillars can protect the machine and ensure the safety of workers [2]. Unstable pillars can increase employee safety risks and potential roof collapse [3]. In addition, as the mining depth increases, the increased ground pressure can also lead to more frequent and serious pillar failures [4]. Therefore, a proper assessment of the stability of the underground pillars is necessary. Assessment of the stability of the existing pillars can provide a reference for the designer to avoid unwanted accidents [5].

In general, pillar stability can be divided into three categories, namely: stable, unstable and failed [6]. Safety Factor (SF) is a common index used in several pillar design methodologies to assess pillar stability in relation to pillar strength and average pillar stress [2]. SF is calculated by dividing the pillar strength by the pillar stress [1]. Theoretically, a rock or coal pillar is considered "unstable" if the SF value is less than 1, and "stable" if it is greater than 1. However, such rigid boundaries are often unreliable, because the occurrence of unstable pillars is also frequently appears when the SF value is above 1 [2][6].

Machine learning techniques have been currently used effectively to evaluate the stability of pillars with better accuracy compared to other methods. This is due to an increase in the availability of pillar stability data [4]. Tawadrous and Katsabanis [7] used Artificial Neural Network (ANN) with logistic activation function and 12 inputs to classify pillar stability. The classification performance showed 90-93% of testing accuracy. Zhou et al. [8] did a comparative study on performances of six supervised learning methods in classifying pillar stability. One of the supervised learning methods compared was ANN with logistic activation function. The highest classification accuracy obtained using ANN was 80.3% with five model inputs. Li et al. [1] proposed Logistic Model Trees (LMT) algorithm to classify pillar stability with five model inputs. The training accuracy and the average accuracy (10-fold CV) of LMT was then compared to other existing algorithm, including ANN. The result showed that the accuracy of LMT model was among the highest with 79.1-80.5% of 10-fold cross validation accuracy. Unfortunately, from the literature research on pillar stability prediction using ANN that has been carried out [1][7][8], the use of the activation function in the ANN algorithm is still limited to the sigmoid (logistic) function. The use of this function can cause vanishing gradient problems when there are many layers in an ANN [9][10]. Vanishing gradient is a situation where the value of the partial derivative of the loss/error function gets closer to zero, and the partial derivative disappears. As a result, ANN will not improve the model weight [11] and ANN performance will not increase. Therefore, the prediction of pillar stability using the ANN algorithm needs to be improved.

One way to overcome the vanishing gradient problem in ANN is to replace the activation function. Several activation functions that have been proposed in recent years to overcome the vanishing gradient problem include ReLU, ELU, and GELU. On the other hand, ensemble learning techniques are also used in ANN to improve ANN performance. Ensemble learning combines the decisions of several predictors in two ways, namely majority vote and averaging. With this combination, the variance of the prediction results will be lower, and consequently the accuracy of the prediction results will increase [12].

In this study, the authors use the South African coal mining data used in [13] by expanding the initial pillar stability categories (intact and failed) into 4 categories based on the stability and its suitability with SF value. In addition, the authors also add a ratio variable referring to research [14]. Next, the authors use Artificial Neural Network-Backpropagation (ANN-BP) with ReLU, ELU, and GELU activation functions for pillar stability classification. Furthermore, classification of pillar stability using ensemble learning with the ANN-BP basic model was also carried out. The results of the classification using ensemble learning are then compared with the performance of a single ANN-BP. The framework of this study can be seen in Fig. 1.

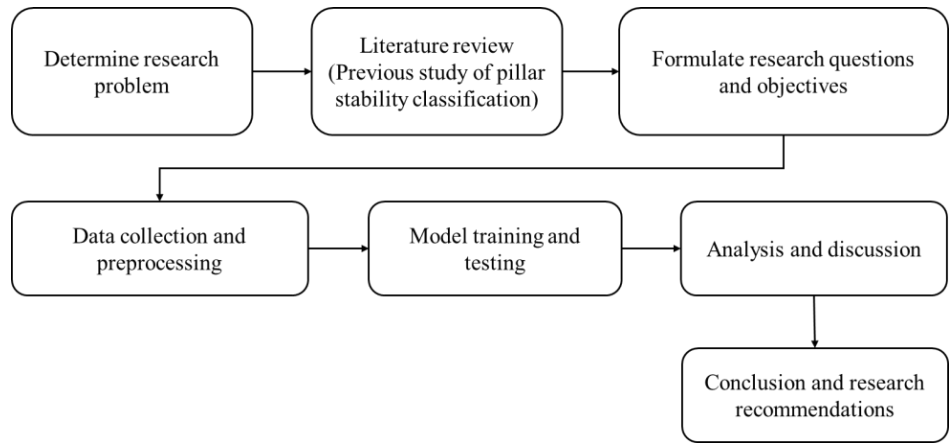

**FIGURE 1.** Research framework

## MATERIALS AND METHOD

### Dataset

The dataset used in this study was taken from a journal written by J.N. van der Merwe and M. Mathey [13] entitled "Update of Coal Pillar Database for South African Coal Mining". The data consists of 423 case histories of coal mines in South Africa with 4 types of variables (depth, mining height, bord with, and pillar width) and 2 types of pillar stability labels (intact and failed). The intact label means that the pillar is stable, while the failed label means that the pillar is unstable which can cause it to collapse. Based on the results of data exploration, it was found that the data contained no missing values, and 337 of the 423 cases in the data were labeled intact. This means that there is a class imbalance in the data. In addition to the 4 variables contained in the data, in this study one additional variable was added, namely Ratio. The value of the Ratio variable is calculated by dividing the pillar width by mining height.

Furthermore, the SF value is also calculated for each case in the data using the equation (1)-(3) [13][15][16].

$$SF = \frac{S_p}{\sigma_p}, \tag{1}$$

$$S_p = 5.47 \frac{w^{0.8}}{h} \; MPa, \tag{2}$$

$$\sigma_p = \frac{25HC^2}{w^2} \; kPa, \tag{3}$$

where $w$ denotes pillar width (meters), $h$ denotes pillar height (meters), 5.47 is the strength of coal material (MPa), $H$ denotes the depth to the bottom of the mine (meters), $C$ denotes the amount of pillar width and the distance between the pillars (bord width) in meters, and 25 is the multiples of the overburden density and gravitation (kPa/m). According to the theory, a rock or coal pillar will be classified as "unstable" if the SF value is less than 1, and "stable" if it is greater than 1. However, the occurrence of unstable or failed pillars is also frequently appears when the SF value is above 1 [1]. Previously, pillar stability was only classified by only looking at its stability in the field. Because of the fact that there are cases of pillar stability that are not in accordance with the theory, in this study an extension of the pillar stability label was carried out in the dataset used by considering its suitability with the safety factor value. Because there are cases of pillar stability that are not in accordance with the theory, in this study an extension of the pillar stability label was carried out in the dataset used by considering its suitability with the safety factor value. The initial labels in the dataset will be expanded from 2 categories (intact and failed) to 4 categories (F0, F1, I0, and I1) with the description of each label shown in Table 1. Because the expanded label contains additional information on the conformity of the pillar stability to the SF, it is necessary to specify the boundaries that determine whether the stability of the pillars is suitable with the calculation of the SF.

1**TABLE 1.** Expanded label descriptions

| Label | Description | Safety Factor (SF) Value |
|---|---|---|
| F0 | Failed with a suitable safety factor | SF value is lower than the specified boundary |
| F1 | Failed without a suitable safety factor | SF value is higher than the specified boundary |
| I0 | Intact with a suitable safety factor | SF value is lower than the specified boundary |
| I1 | Intact without a suitable safety factor | SF value is higher than the specified boundary |

In this study, cases with class labels F1 and I1 are considered as outliers in the SF data for each label (intact and failed). Referring to Yang et al [17], assuming the data is normally distributed, the general equation for calculating the outlier threshold in a data is presented in equations (4) and (5).

$$T_{min} = mean - a \times SD, \tag{4}$$

$$T_{max} = mean + a \times SD, \tag{5}$$

where $T_{min}$ and $T_{max}$ denote the minimum and maximum threshold respectively. The mean denotes the average of the data, and SD denotes the standard deviation of the data. Thus, the boundary is determined by calculating SF threshold for each label (intact and failed). The threshold used in this study was calculated from the average and standard deviation of the safety factor for each label using equations (4) and (5) with the value $a = 1$. The threshold used for each label is as follows.

$$T_{failed} = \text{Avg. SF Failed} + \text{Std. SF Failed} = 2.48, \quad (6)$$

$$T_{intact} = \text{Avg. SF Intact} + \text{Std. SF Intact} = 1.42, \quad (7)$$

where $T_{failed}$ denotes the maximum SF value limit is said to be suitable for failed labels and $T_{intact}$ denotes the minimum SF value limit is said to be suitable for intact labels. The labeling rules are presented in Table 2.

2**TABLE 2.** Labeling conditions

| Conditions | Label |
|---|---|
| *Failed & SF ≤ 2.48* | F0 |
| *Failed & SF > 2.48* | F1 |
| *Intact & SF ≥ 1.42* | I0 |
| *Intact & SF < 1.42* | I1 |

## Preprocessing

Before conducting model training, preprocessing is carried out on the dataset used. This step is needed so that the data can be used as input to the model and the model can learn the characteristics of the dataset better. The preprocessing performed on the dataset includes three stages, namely oversampling, label encoding, and split dataset. The oversampling method used in this study is SMOTE. After oversampling, the number of cases for each label was 312. The difference in the number of cases for each category before and after SMOTE can be seen in table 3. Next, label encoding was also performed for each class. This needs to be done because ANN-BP cannot read categorical data types. Class F0 is encoded as 0, class F1 is encoded as 1, class I0 is encoded as 2, and class I1 is encoded as 3. Furthermore, the data is finally divided into training data sets, validation data, and testing data. There are four combinations of data proportions used to test each model, each of which can be seen in Table 4.

**TABLE 3.** The difference in the number of cases for each stability category before and after SMOTE

| Category | Number of Cases (Before SMOTE) | Number of Cases (After SMOTE) |
|---|---|---|
| F0 | 70 | 312 |
| F1 | 16 | 312 |
| I0 | 312 | 312 |
| I1 | 25 | 312 |

**TABLE 4.** Types of proportions and percentages of training, validation, and testing data

| Data Proportion | % Training, Validation, and Testing |
|---|---|
| 1 | 80% Training, 20% Testing |
| 2 | 70% Training, 30% Testing |
| 3 | 80% Training, 10% Validation, 10% Testing |
| 4 | 70% Training, 15% Validation, 15% Testing |

## Model Training

The ANN-BP architecture used in this study is multilayer perceptron (MLP). The number of neurons used in the input layer and the output layer in this architecture are 5 and 4 respectively (because there are 5 inputs, namely depth, pillar width, mining height, bord width, and ratio, and 4 output labels, namely F0, F1, I0, and I1). In this study, it was determined that the number of hidden layers was 4 and the number of neurons contained in the hidden layer were 512, 256, 256, and 128 respectively. There were 3 ANN-BP models used in this study which were differentiated based on the type of activation function used, namely ReLU [18], ELU [19], and GELU [9].

Mathematically, the output of the ReLU activation function can be expressed as follows [18].

$$ReLU(x) = \max(0, x), x \in \mathbb{R}. \tag{8}$$

From the mathematical expression of ReLU, we can see that not all neurons are activated in ReLU function. This is why ReLU function is said to be more effective than other activation functions. However, ReLU function cannot produce negative values. This can lead to particular case of vanishing gradient problem, which is dying ReLU problem. On the other hand, ELU activation function was proposed as an improvement of ReLU because it can produce negative values. Mathematically, the output of the ELU activation function can be expressed as follows [19].

$$ELU(x) = \max(0, x) + \min(0, \alpha(e^x - 1)), x \in \mathbb{R}, \tag{9}$$

where $\alpha$ is hyperparameter that controls the output of negative inputs. GELU activation function is a State of the Art of activation function proposed in 2020 by Hendrycks & Gimpel. While the deactivation of neurons in ReLU and ELU function is still stochastic and highly dependent on input values, GELU is formulated to deactivate neurons deterministically. GELU assumes normally distributed inputs with mean 0 and standard deviation 1 ($X \sim N(0,1)$), and uses the standard Gaussian cumulative distribution function ($\Phi(x)$) as the input multiplier. GELU function is defined as follows [9].

$$GELU(x) = xP(X \leq x) = x\,\Phi(x) = x \cdot \frac{1}{2}\left[1 + erf\left(x/\sqrt{2}\right)\right] \tag{10}$$

The type of ANN-BP ensemble learning used in this study is bagging. The ensemble learning process begins with bootstrapping, or sampling with replacement from the training data to serve as new training data for each basic model of ensemble learning (ANN-BP ReLU, ANN-BP ELU, and ANN-BP GELU). There are 3 types of bootstrap percentages used in this study, they are 70%, 80%, and 90%. After bootstrapping, the training process for each ANN-BP is carried out independently.

The process of determining the class in ensemble learning is done by doing a majority vote on the prediction results of each basic model. If the prediction results of the three ANN-BP are different, then the class that becomes the final prediction for ensemble learning is the class with the smallest encoding label. In this study, class with the smallest encoding label is F0. Class F0 is the best choice when there are differences in predictions among the three basic models, because class F0 is a class with the smallest risk of prediction error compared to the other classes (F1, I0, and I1). Class F0 is said to have the smallest risk of prediction error because the stability of the pillars with class F0 is in accordance with the predicted stability based on the calculation of the safety factor. In addition, because the predicted stability in class F0 is failed, the possibility of the pillar being built is also lower, and consequently the operational costs incurred are also smaller. The majority vote and the ensemble learning process carried out in this study is shown in Table 5 and Fig. 2 respectively.

**TABLE 5.** Example of Ensemble Learning Decision Making Process using Majority Vote

| Predicted Class by ANN-BP ReLU | Predicted Class by ANN-BP ELU | Predicted Class by ANN-BP GELU | Final Prediction by Ensemble Learning |
|---|---|---|---|
| 1 | 1 | 1 | 1 |
| 2 | 2 | 1 | 2 |
| 1 | 0 | 2 | 0 |
| 0 | 3 | 1 | 0 |

In this study, the model simulation was implemented using the Python programming language and executed on Google Colab with GPU running time. The model is built using the TensorFlow library and trained for 10 times. In TensorFlow, the batch_size parameter specifies the number of samples in a batch that are inputted into the neural network before updating the model parameters, while the epochs parameter specifies the number of times the entire dataset is inputted into the neural network. In this simulation the batch_size used for each model is 16 with the maximum number of epochs for one training is 400. Models without data validation are trained using accuracy Early Stopping with a value of patience = 10, which means the model will stop the training process if the training accuracy value is not improved after 10 epochs. Meanwhile, models with validation data are trained using val_loss Early Stopping with a value of patience = 10, which means the model will stop the training process if the validation loss value does not improve after 10 epochs.

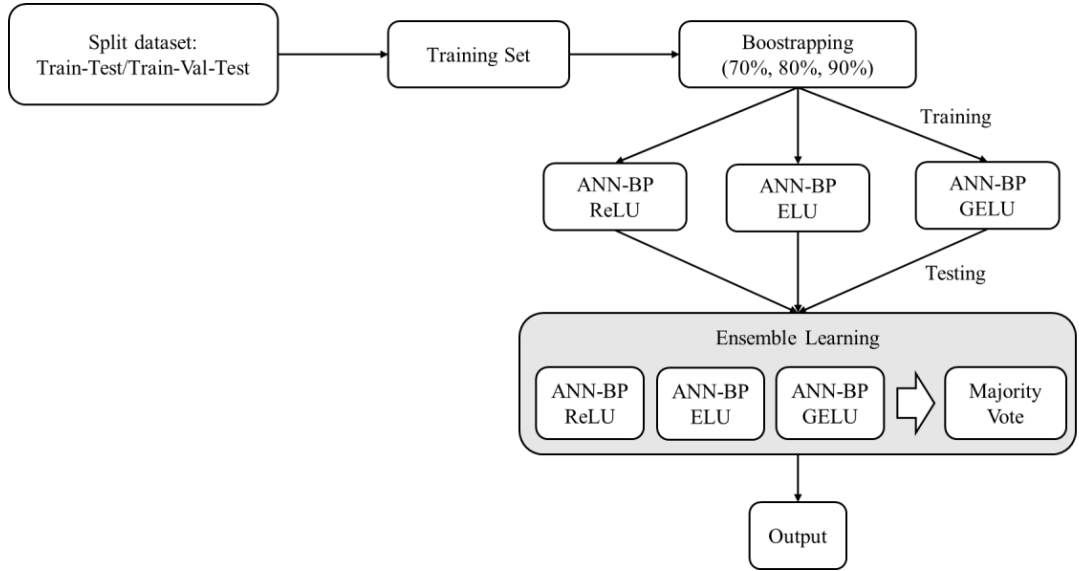

**FIGURE 2.** Ensemble Learning Scheme on ANN-BP ReLU, ANN-BP ELU, and ANN-BP GELU

# RESULTS AND DISCUSSION

The average accuracy and standard deviation of accuracy from 10 trials of the four models (ANN-BP ReLU, ANN-BP ELU, ANN-BP GELU, and Ensemble Learning) are presented in Table 6-7.

**TABLE 6.** Average accuracy of the four models

| Data Proportion | ANN-BP ReLU | ANN-BP ELU | ANN-BP GELU | Ensemble Learning (70%) | Ensemble Learning (80%) | Ensemble Learning (90%) |
|---|---|---|---|---|---|---|
| 1 | 83.48% | 82.48% | 86.32% | 88.32% | 88.68% | 88.76% |
| 2 | 83.60% | 85.44% | 87.73% | 90.64% | 90.37% | 90.21% |
| 3 | **82.16%** | 82.56% | **83.52%** | 87.12% | 87.92% | 86.48% |
| 4 | 85.48% | 83.24% | **87.98%** | 89.15% | 89.20% | **89.31%** |

Based on the results of the average model accuracy, ANN-BP with GELU activation function produces a better average accuracy compared to ANN-BP with ReLU and ELU activation functions for each data proportion, which is at least 1.36% higher in the 3$^{rd}$ type of data proportion. The best average accuracy produced by ANN-BP GELU is obtained in the 4$^{th}$ data proportion, which is 87.98%. On the other hand, the use ensemble learning can increase the average accuracy produced by a single ANN-BP, at least by 1.17% in the 4$^{th}$ type data proportion. The best average accuracy produced using ensemble learning is obtained using 70% percentage of bootstrap in the 2$^{nd}$ type data proportion, which is 90.64%. This means that the model can provide the right pillar stability predictions for around 90.64% of the data from all the testing data used, or in other words, the chance that the model gives the right predictions for the testing data used is 90.64%.

Overall, ensemble learning increases the average accuracy produced using single ANN-BP for each data proportion. However, the difference of average accuracy in ensemble learning for each bootstrap percentages is still very small (not reaching 1%).

TABLE 7. Standard deviation of model accuracy

| Data Proportion | ANN-BP ReLU | ANN-BP ELU | ANN-BP GELU | Ensemble Learning (70%) | Ensemble Learning (80%) | Ensemble Learning (90%) |
|---|---|---|---|---|---|---|
| 1 | 1.83% | **6.84%** | 3.65% | 0.75% | 0.98% | 1.18% |
| 2 | 2.32% | 3.42% | 1.71% | 0.98% | 0.61% | 0.78% |
| 3 | 2.30% | 2.92% | 3.83% | 1.66% | **1.86%** | 1.53% |
| 4 | 2.31% | 4.62% | **1.57%** | 1.55% | 0.97% | **0.47%** |

Based on the results from Table 7, the standard deviation of ensemble learning accuracy only ranges from 0.47-1.86%, with the smallest standard deviation of accuracy obtained using Ensemble Learning 90% in the 4[th] data proportion and the largest standard deviation of accuracy obtained using Ensemble Learning 80% in 3[rd] data proportion. Meanwhile, the standard deviation of single ANN-BP accuracy ranges from 1.57-6.84%, with the smallest standard deviation of accuracy obtained using the GELU activation function in the 4[th] data proportion and the largest standard deviation of accuracy obtained using the ELU activation function in the 1[st] data proportion.

Because the value is smaller, the standard deviation of ensemble learning accuracy is said to be better when compared to the standard deviation of single ANN-BP accuracy. This means that the percentage accuracy of the prediction results using ensemble learning does not fluctuate too much from the average accuracy.

In addition to the average accuracy and standard deviation of accuracy, the $F_1$ score for each class label is also calculated and presented in Table 8-15. In particular, cases with class label F1 must be prioritized compared to cases with other class labels (F0, I0, and I1) because F1 is the most dangerous class. It is said to be the most dangerous because based on the calculation of the safety factor, pillars in the F1 class are categorized as intact, but in reality these pillars failed. Cases in this class will cause greater operational losses compared to other classes. Therefore, the authors assume that it is much worse to miss a failed stability prediction than to give a false alarm for an intact pillar stability. This means, in the F1 category, recall in more important than precision. In this study, F1 class recall is considered to be twice as important as F1 class precision. Therefore, the $F_2$ score for each model is calculated for the F1 category by choosing the value $\beta = 2$ in the $F_\beta$ evaluation metric function. The results of calculating the $F_1$ score and $F_2$ score are presented in Table 8-11 and Fig 3 respectively.

TABLE 8. $F_1$ score for 1[st] data proportion

| Label | ANN-BP ReLU | ANN-BP ELU | ANN-BP GELU | Ensemble Learning (70%) | Ensemble Learning (80%) | Ensemble Learning (90%) |
|---|---|---|---|---|---|---|
| F0 | 72.83% | 74.46% | 78.60% | 80.52% | 81.37% | 81.98% |
| F1 | 94.91% | 95.49% | 96.61% | 97.06% | 97.13% | 97.42% |
| I0 | 80.91% | 81.98% | 84.51% | 86.59% | 87.48% | 87.94% |
| I1 | 83.58% | 76.91% | 84.73% | 88.28% | 88.16% | 87.37% |

TABLE 9. $F_1$ score for 2[nd] data proportion

| Label | ANN-BP ReLU | ANN-BP ELU | ANN-BP GELU | Ensemble Learning (70%) | Ensemble Learning (80%) | Ensemble Learning (90%) |
|---|---|---|---|---|---|---|
| F0 | 72.57% | 74.54% | 79.27% | 82.96% | 82.33% | 82.16% |
| F1 | 95.58% | 96.16% | 96.84% | 98.19% | 97.99% | 97.94% |
| I0 | 82.72% | 85.12% | 86.53% | 91.01% | 91.70% | 91.45% |
| I1 | 82.82% | 85.19% | 87.66% | 90.24% | 89.54% | 89.42% |

TABLE 10. $F_1$ score for 3[rd] data proportion

| Label | ANN-BP ReLU | ANN-BP ELU | ANN-BP GELU | Ensemble Learning (70%) | Ensemble Learning (80%) | Ensemble Learning (90%) |
|---|---|---|---|---|---|---|
| F0 | 70.15% | 70.56% | 74.14% | 77.55% | 77.62% | 74.68% |
| F1 | 94.85% | 94.23% | 93.87% | 96.68% | 97.37% | 96.68% |
| I0 | 79.28% | 79.77% | **76.52%** | 88.05% | 88.93% | 86.44% |
| I1 | 79.78% | 81.33% | 84.53% | 83.84% | 84.91% | 84.42% |

TABLE 11. $F_1$ score for 4th data proportion

| Label | ANN-BP ReLU | ANN-BP ELU | ANN-BP GELU | Ensemble Learning (70%) | Ensemble Learning (80%) | Ensemble Learning (90%) |
|---|---|---|---|---|---|---|
| F0 | 70.15% | 70.56% | 74.14% | 77.55% | 77.62% | 74.68% |
| F1 | 94.85% | 94.23% | 93.87% | 96.68% | 97.37% | 96.68% |
| I0 | 79.28% | 79.77% | **76.52%** | 88.05% | 88.93% | 86.44% |
| I1 | 79.78% | 81.33% | 84.53% | 83.84% | 84.91% | 84.42% |

Based on the $F_1$ score obtained, among the other three types of activation function, GELU activation function provides better model performance in classifying the majority of class labels in each type of data proportion (except data with label I1 in the 1st, 2nd and 3rd data proportion, and data with label I0 in the 3rd data proportion). The best $F_1$ score obtained using GELU activation function can be found in the 2nd data proportion, with the $F_1$ score from class F0 is 79.27%, class F1 is 96.84%, class I0 is 86.53%, and class I1 is 87.66%. In this data, ensemble learning provides equal or better performance for detecting the stability of each label for almost every proportion of data. The best $F_1$ score obtained using ensemble learning can be found in the 2nd data proportion. The bootstrap percentage that was used is 70%, with the $F_1$ score from class F0 is 82.96%, class F1 is 98.19%, class I0 is 91.70%, and class I1 is 90.24%.

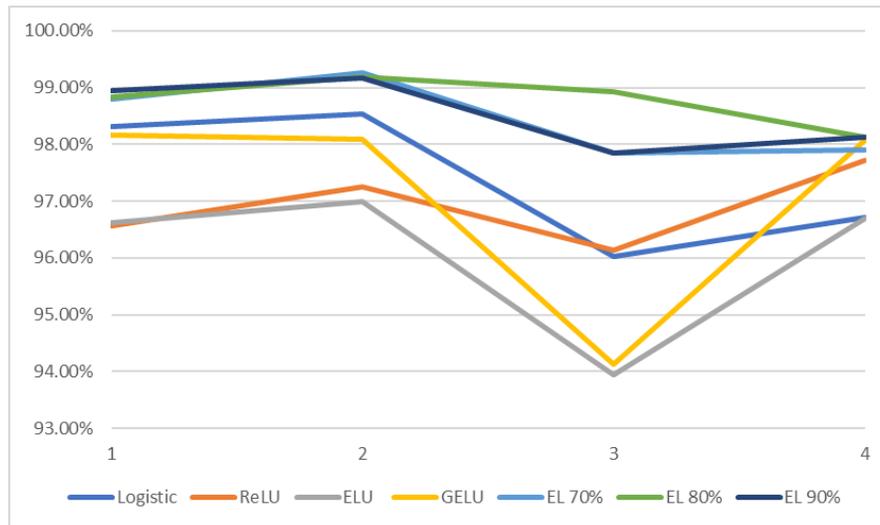

**FIGURE 3.** $F_2$ Score Comparison of Single ANN-BP and Ensemble Learning Model

Based on Fig. 3, it can be seen that ensemble learning provides better performance in detecting class F1 for all types of data proportions compared to single ANN-BP. The $F_2$ score of class F1 obtained using ensemble learning reaches 99.27% (Ensemble Learning 70%). This means that the ensemble learning model has a very good ability in focus on detecting the presence of class F1.

## CONCLUSION

Among the three types of ANN-BP activation functions, GELU activation function provides the best performance in classifying pillar stability measured based on average accuracy, standard deviation, and $F_1$ score. This might be caused by its ability to deterministically deactivate neurons in the neural network during the training process. The best average accuracy using ANN-BP GELU is 87.98% (in the 4th data proportion) and the best standard deviation of accuracy obtained using ANN-BP GELU is 1.57% (4th data proportion).

Ensemble learning (EL) provides excellent performance in the pillar stability classification measured by accuracy, standard deviation, $F_1$ score. The best average accuracy using EL is 90.64% (EL 70%) and the best standard deviation of accuracy using EL is 0.47% (EL 90%). Ensemble learning can also improve the performance of the pillar stability classification of a single ANN-BP, with an increase in average accuracy of at least 1.17% and reduce the standard deviation of accuracy to a maximum of 1.86%.

Finally, the change of the activation functions in the previous ANN algorithm can improve the clasification results. It gives the same or better classification performance with less number of inputs. The additional implementation of ensemble learning also stabilize the classification process by making the classification results more robust against changes of input.

## ACKNOWLEDGMENTS

The research is supported by Hibah Riset Penugasan FMIPA UI No. 003/UN2.F3.D/PPM.00.02/2022